\documentclass[accepted,specialissue]{melba}

\usepackage{amsmath,amsfonts}
\usepackage{booktabs}
\usepackage{tabularx}
\usepackage{array}
\usepackage{multirow}
\usepackage{makecell}
\usepackage{url}
\usepackage{graphicx}
\usepackage{placeins}
\usepackage{stfloats}
\usepackage{makecell}

\newcommand{\resource}{GLI-AL}

\melbaid{}
\doi{}
\melbaauthors{Xingyu Xiang, Shuang Hao, Fan Wang, Jianhua Ma, and Chunfeng Lian}
\email{chunfeng.lian@xjtu.edu.cn}
\volume{2026}
\firstpageno{1}
\melbayear{2026}
\datesubmitted{}
\datepublished{}

\melbaspecialissue{MICCAI Open Data 2026 x MELBA}
\melbaspecialissueeditors{}

\ShortHeadings{\resource}{Xiang et al.}

\title{\resource: A Multi-Modal Glioma MRI Label Resource with Unified Anatomy-Lesion Labels}

\author{
	\name Xingyu Xiang\aff{2,3},
	\name Shuang Hao\aff{1,3},
	\name Fan Wang\aff{1,3},
	\name Jianhua Ma\aff{1,3},
	\name Chunfeng Lian\aff{2,3}
}
\affiliations{%
	\num 1 \addr Key Laboratory of Biomedical Information Engineering of Ministry of Education, School of Life Science and Technology, Xi'an Jiaotong University, Xi'an, China \\
	\num 2 \addr School of Mathematics and Statistics, Xi'an Jiaotong University, Xi'an, China \\
	\num 3 \addr Research Center for Intelligent Medical Equipment and Devices (IMED), Xi'an Jiaotong University, Xi'an 710049, China
}

\abstract{Existing BraTS-GLI datasets provide a widely used benchmark for adult glioma MRI segmentation, but their task definition focuses on tumor subregions and does not systematically represent coexisting white matter hyperintensities (WMH). In joint segmentation settings, such unlabeled abnormalities introduce task-specific label noise by treating pathological regions as normal tissue. To address this limitation, we introduce BraTS-GLI Anatomy-Lesion, a controlled-access, labels-only derived resource built from the BraTS 2023-GLI training cohort. The resource provides 1,251 unified eight-class anatomy-lesion label sets aligned with the original four-modal MRI cases, including image-repair labels for 116 cases requiring repaired imaging inputs. The cohort is organized into a 394-case purified subset and an 857-case extended subset, with case-level metadata covering label source, image-repair requirements, quality-control status, access conditions, checksums, and release boundaries. Compared with the original BraTS-GLI annotations, the resource substantially expands foreground supervision by incorporating healthy brain tissues and previously unlabeled coexisting abnormalities within a unified label space. A validation study using MedNeXt and T1/FLAIR inputs suggests that WMH-aware supervision preserves healthy-tissue segmentation performance across both in-domain GLI and external WMH datasets, while improving sensitivity to coexisting lesions relative to noisy-control training. The resource is intended for scientific research and supports joint anatomy-lesion supervision, label-noise analysis, and reproducible evaluation. Data are available at \url{https://www.synapse.org/Synapse:syn75210889/wiki/}, and code is available at \url{https://github.com/xyx200/brats-gli-anatomy-lesion-code}. The data resource DOI is \url{https://doi.org/10.7303/SYN75210889}.
}


\begin{document}
	\emergencystretch=2em
	\raggedbottom
	
	\twocolumn[\maketitle]
	
	\section{Background}
	\enluminure{B}{raTS} datasets provide multi-center, pre-operative, multi-parametric MRI and expert tumor-subregion annotations for brain tumor segmentation research, and they are among the central public benchmarks for machine learning in glioma imaging~\citep{menze2015multimodal,bakas2017tcga,bakas2017tcgagbm,bakas2017tcgalgg,baid2021rsna,karargyris2023medperf}. BraTS 2023-GLI contains four-modal images of adult glioma cases, with tumor annotations covering necrotic tumor core (NCR), peritumoral edema (ED), and enhancing tumor (ET) regions~\citep{baid2021rsna}. This task definition is highly effective for glioma segmentation, but it also raises an issue that is easily overlooked in joint segmentation settings: datasets usually annotate only the target lesion and do not systematically annotate other abnormalities that may coexist in the patient's brain~\citep{rudie2019multi,baid2021rsna}.
	
	WMH is a common coexisting abnormality in BraTS-GLI brain MRI. Rudie et al. added expert WMH annotations to 285 BraTS 2018 training cases and reported that 68.8\% of the cases contained at least 100 mm$^3$ of WMH~\citep{rudie2019multi}. This finding shows that unlabeled WMH in glioma cohorts is not an isolated occurrence. In joint segmentation settings, these unlabeled WMH regions are implicitly treated as healthy tissue during training. Consequently, the resulting supervision signal may encourage models to misclassify pathological regions as normal anatomy.
	
	When the same voxel space needs to represent healthy brain tissues, tumor lesions, and coexisting abnormalities at the same time, the original BraTS-GLI tumor-subregion labels cannot directly serve as a joint supervision target. This resource gap appears at two levels: coexisting WMH is not systematically represented in the original task, and healthy tissue structures are also not included in the same label space. Table~\ref{tab:resource-comparison} summarizes the data-resource context: established anatomy tools can produce healthy-tissue labels~\citep{fischl2012freesurfer,billot2023synthseg}, and major brain lesion datasets provide single-type tumor or stroke lesion annotations~\citep{BraTS2023SynapsePage,absher2024soop}, but they do not provide healthy-tissue labels, coexisting-lesion labels, and auditable provenance records together in a single reusable data resource. Examples of T1/FLAIR, original BraTS labels, added lesion labels, and unified healthy tissue--lesion labels are shown in Figure~\ref{fig:label-example}. To address this gap, we introduce \resource, a WMH-aware anatomy–lesion label resource built upon the full BraTS 2023-GLI training cohort. Specifically, we (1) establish WMH-aware stratification across all cases and distinguish purified and extended subsets, (2) construct a unified label space covering both healthy tissues and lesion structures, and (3) provide traceable label-source metadata and quality-control records for all released cases and files. The resulting resource enables joint supervision of normal anatomy and pathological structures, supports label-noise sensitivity analysis, and facilitates case-source-stratified experimental design. To our knowledge, no publicly available BraTS-GLI-derived resource currently integrates healthy-tissue labels, tumor annotations, WMH-aware stratification, and auditable provenance records within a unified label space.

	\begin{table*}[!t]
		\centering
		\caption{Resource-context comparison}
		\label{tab:resource-comparison}
		\footnotesize
		\setlength{\tabcolsep}{3pt}
		\renewcommand{\arraystretch}{1.12}
		\renewcommand{\tabularxcolumn}[1]{m{#1}}
		\begin{tabularx}{\textwidth}{@{}
				>{\raggedright\arraybackslash}m{0.32\textwidth}
				>{\centering\arraybackslash}X
				>{\centering\arraybackslash}X
				>{\centering\arraybackslash}X
				>{\centering\arraybackslash}X
				@{}}
			\toprule
			\multicolumn{1}{@{}>{\centering\arraybackslash}m{0.32\textwidth}}{Resource} &
			\makecell[c]{Anatomy\\labels} &
			\makecell[c]{Primary target\\lesion labels} &
			\makecell[c]{Comorbid\\lesion labels} &
			\makecell[c]{Case\\stratification} \\
			\midrule
			FreeSurfer/SynthSeg &
			\checkmark &
			$\times$ &
			$\times$ &
			$\times$ \\
			BraTS &
			$\times$ &
			\checkmark &
			$\times$ &
			$\times$ \\
			SOOP stroke dataset &
			$\times$ &
			\checkmark &
			$\times$ &
			$\times$ \\
			\resource &
			\checkmark &
			\checkmark &
			\checkmark &
			\checkmark \\
			\bottomrule
		\end{tabularx}
	\end{table*}

	\section{Summary}
	This resource is intended for research on medical image joint segmentation, brain tumor MRI, label noise, and AI-ready datasets. It provides WMH-aware BraTS-GLI derived labels with transparent label sources and case-level metadata. Researchers can conduct joint training and evaluation for healthy tissues and lesions within the same label space, as well as sensitivity analyses stratified by label source and case subset.
	
	The resource corresponds to 1251 four-modal MRI cases from the BraTS 2023-GLI training set. This paper releases only derived labels and does not redistribute MRI images. The labels use NIfTI format and remain aligned with the upstream BraTS case-folder layout, inheriting the upstream standard-space settings: SRI24 template space, 1 mm$^3$ resolution, 240$\times$240$\times$155 voxel dimensions, and skull-stripped images~\citep{baid2021rsna}. The integer codes of the unified labels are: 0 for background, 1 for cortical gray matter (GM), 2 for basal ganglia (BG), 3 for white matter (WM), 4 for Lesion, 5 for ventricle (Ven), 6 for cerebellum (Cer), and 7 for brainstem (BS). This coding places lesion labels and healthy tissue structures into a single supervision target, so downstream models no longer need to switch between incompatible label definitions. Combined with the upstream BraTS masks, users can retain NCR, ED, and ET, and identify voxels in the unified Lesion class that lie outside the upstream whole-tumor mask as the newly added lesion component. The MedNeXt use case in the validation experiments~\citep{isensee2021nnunet,roy2023mednext} uses only T1 and FLAIR; this is an experimental setting, not a modality limitation of the label resource.
	
	\begin{figure*}[!t]
		\centering
		\setlength{\tabcolsep}{3pt}
		\renewcommand{\arraystretch}{0.92}
		\begin{tabular}{@{}*{5}{>{\centering\arraybackslash}m{0.19\textwidth}}@{}}
			\textbf{T1} & \textbf{FLAIR} & \textbf{BraTS label} & \textbf{Added lesion} & \textbf{Unified label} \\
			\includegraphics[width=\linewidth]{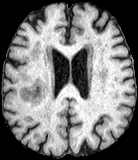}
			& \includegraphics[width=\linewidth]{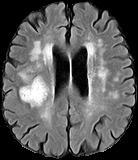}
			& \includegraphics[width=\linewidth]{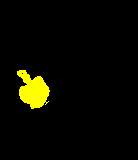}
			& \includegraphics[width=\linewidth]{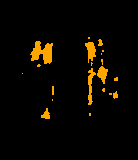}
			& \includegraphics[width=\linewidth]{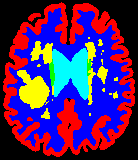}
		\end{tabular}
		\caption{Example images and derived labels. The BraTS label image displays the original BraTS tumor subregions merged as a whole-tumor mask. Added lesion denotes a lesion region that is visible on FLAIR, not covered by the original BraTS tumor-task label, and included in the Lesion class in this paper. The unified healthy tissue--lesion label supplements healthy tissue labels for cortical gray matter, basal ganglia, white matter, ventricle, cerebellum, and brainstem.}
		\label{fig:label-example}
	\end{figure*}
	
	The full training-set resource is divided into a 394-case purified subset and an 857-case extended subset; the construction process and label sources are described in Methods. The purified subset includes cases with expert-negative WMH conclusions from the Rudie et al. expert WMH annotation resource~\citep{rudie2019multi}, together with additional WMH-negative cases identified by model screening. The extended subset merges additional lesion constraints from the remaining cases into the unified Lesion class. In addition to the unified labels, 116 cases also provide repair labels with two foreground classes, representing whole tumor and WMH, respectively. These repair labels copy the label resource of Rudie et al.~\citep{rudie2019multi}. The repair-label cases contain 105,417 WMH repair voxels in total, corresponding to 0.89\% of their accompanying whole-tumor foreground volume, with a median repaired WMH volume of 448.5 voxels per case. These repair labels allow users, after obtaining the upstream BraTS MRI, to reproduce the repaired image inputs used in this study with the accompanying code. Users need to pair this resource's labels with upstream BraTS 2023-GLI MRI. Apart from Synapse CLI, Python, NiBabel/SimpleITK, or equivalent medical-image I/O tools, no additional registration or resampling is required.
	
	To quantify the amount of information added by this resource, we compared each released unified label with the corresponding original BraTS-GLI expert lesion mask after mapping both files to the same case identifier. The original mask was treated as the pre-existing tumor-task foreground. In contrast, the released unified label contains both healthy-tissue structures and the final Lesion class. This comparison shows that the resource adds foreground labels outside the original expert lesion mask for all 1251 cases. These added labels occupy 1.51 billion voxels, equal to 92.7\% of the released foreground. Most of this added foreground comes from healthy-tissue structures that were absent from the original tumor-task label. The lesion-specific addition is smaller but directly addresses the label-noise problem: 857 cases contain additional Lesion voxels outside the original expert lesion foreground, totaling 2.20 million voxels. These voxels represent 1.8\% of the final Lesion class and mark coexisting lesion constraints that would otherwise be treated as healthy tissue in a joint segmentation target.
	
	From the perspective of AI-readiness and FAIR use, version v1.0.0 provides controlled-access derived labels and metadata through Synapse project ID \texttt{syn75210889}. Labels are released as NIfTI files in BraTS-compatible case folders, and case identifiers can be aligned with upstream BraTS 2023-GLI MRI. The release metadata record the stratification composition, case-level provenance and QC status, internal-ID mapping for probability-map generation and label fusion, access conditions, file sizes, and SHA-256 digests; the repository also provides a Synapse CLI download entry point. These metadata allow users to distinguish case sources, label sources, and release boundaries in training, evaluation, and manuscript reporting, rather than treating all derived labels as homogeneous samples. Users must first obtain access to the upstream BraTS project \texttt{syn51156910}, then apply for access to \texttt{syn75210889} through the email workflow in this resource's wiki and accept the terms for the derived resource.

	\section{Discussion}
	\subsection{Strengths}
	The main strength of this resource is that it handles lesion-label noise caused by unlabeled WMH in BraTS-GLI at the data-source level while keeping the resulting decisions auditable. Label-noise identification, purified/extended stratification, an image-repair reproduction entry point, case-level provenance, QC status, and file-checksum records are fixed parts of the release. This organization allows users to select cases by source, reproduce release boundaries, and report training or evaluation data use without treating all derived labels as homogeneous samples.
	
	A second strength is that the resource provides unified eight-class anatomy-lesion labels for the full training set, allowing lesions and healthy anatomical structures to be jointly modeled within one training target. This makes the resource suitable for joint healthy tissue--lesion supervision and label-noise sensitivity analyses without switching between incompatible task labels.
	
	\subsection{Known limitations}
	The purified subset is not equivalent to 394 cases that have all undergone new manual WMH annotation. It consists of both expert-negative samples and model-screened samples, so users should report its screening source. The additional lesion constraints for the extended subset come from the fusion of automatic tools DeepWMH~\citep{li2024deepwmh} and LST-AI~\citep{wiltgen2024lstai}, which may introduce tool bias, missed detections, and limited sensitivity to small lesions. Healthy-tissue labels come from the automatic tool TumorSynth~\citep{wu2026tumorsynth} and a probabilistic fusion strategy, and they are not equivalent to manual whole-brain anatomical annotations.
	
	This resource systematically handles WMH comorbid lesions; other coexisting abnormalities, such as microbleeds and lacunar infarcts, have not yet been systematically completed. The external WMH validation set contains only T1 and FLAIR, so the use case in this paper cannot prove the value of T1ce and T2 in all downstream tasks.
	
	\subsection{Responsible use}
	\resource{} is intended only for scientific research and should not be used for direct clinical diagnosis, treatment decisions, or individual risk assessment. Users should report the data tier and label source used, and should state the limitations of automatically generated labels. Any re-identification attempt is inconsistent with the intended use of this resource. If errors are found in the derived labels, version records, or accompanying code, users should contact the resource contact for this paper: \texttt{xxy200200@stu.xjtu.edu.cn}; issues involving original data or upstream terms should be directed to the official BraTS contact. Future versions will investigate extension of joint healthy tissue and lesion labels to more cases.
	
	\section{Resource Availability}
	\subsection{Summary statement}
	This resource provides BraTS-GLI derived labels and case-level metadata for joint healthy tissue--lesion supervision, addressing the problem that the original tumor-subregion labels cannot directly represent the relationship between coexisting WMH and healthy tissue structures. The resource scope and access route are described in Summary and Resource Availability, the stratified construction workflow is described in Methods, and quality control and example validation results are described in Validation.
	
	\subsection{Data and code location}
	The Synapse repository for this resource is \url{https://www.synapse.org/Synapse:syn75210889/wiki/}, and the Synapse project ID is \texttt{syn75210889}. The data resource DOI is \url{https://doi.org/10.7303/SYN75210889}. The accompanying code repository is \url{https://github.com/xyx200/brats-gli-anatomy-lesion-code}, which releases the processing scripts authored by this study and the MedNeXt/nnUNet unified-label evaluation adaptation files. This code repository does not redistribute BraTS imaging data, MedNeXt model or training code, model weights, DeepWMH, LST-AI, TumorSynth, NiftySeg, or a complete third-party software environment; third-party tools or model implementations used in this workflow should be obtained from their official sources. Resource versions start from v1.0.0; future label or metadata changes will be documented through versioned records.
	
	This derived resource is a controlled-access labels-only resource. Users must first obtain upstream BraTS 2023 data access in the official BraTS 2023 Synapse project \texttt{syn51156910}~\citep{BraTS2023SynapsePage} and accept the applicable terms, then apply for access to \resource{} through the email template in the \texttt{syn75210889} wiki and accept the terms for this derived resource.
	
	\subsection{Potential use cases}
	This resource is intended only for scientific research. Expected use cases include glioma MRI joint segmentation research under implicit label noise; joint modeling of healthy brain tissues and unified lesions; analysis of training-set cleaning and the impact of label noise; provenance-based stratified training; and, when shared MRI modalities are available, cross-resource validation from the GLI cohort to external research data. These use cases depend on joint labels for healthy tissues and lesions in the same space and cannot be directly completed with the original BraTS tumor-subregion labels.
	
	\subsection{Licensing}
	This resource is built from controlled-access BraTS 2023-GLI data. Users must first complete the data-access application on the official BraTS 2023 Synapse page and obtain access to the original BraTS-GLI data; this derived resource does not replace or bypass that process. This resource is released under \textit{Creative Commons Attribution-NonCommercial 4.0 International} (CC BY-NC 4.0, \url{https://creativecommons.org/licenses/by-nc/4.0/}), consistent with the upstream non-commercial use conditions, and remains subject to the applicable BraTS 2023 post-Challenge Terms and Conditions; \texttt{LICENSE.txt} is provided in the repository root.
	
	Publications using this resource must comply with the official BraTS citation requirements and retain the data-source statement: ``Data used in this publication were obtained as part of the Brain Tumor Segmentation (BraTS) Challenge project through Synapse ID: syn51156910.'' Processing scripts authored by this study in the accompanying code repository are released under the MIT License. Unified-label evaluation files adapted from the MedNeXt/nnUNet evaluation package are released under the Apache License, Version 2.0, with upstream copyright notices and descriptions of this study's modifications retained. This paper uses MedNeXt as a validation model and does not claim to modify or redistribute its model architecture, training implementation, or model weights. Users must not use this resource for direct clinical application, re-identification attempts, or sharing downloaded label files with unapproved individuals.
	
		\begin{figure*}[t]
		\centering
		\includegraphics[width=\textwidth]{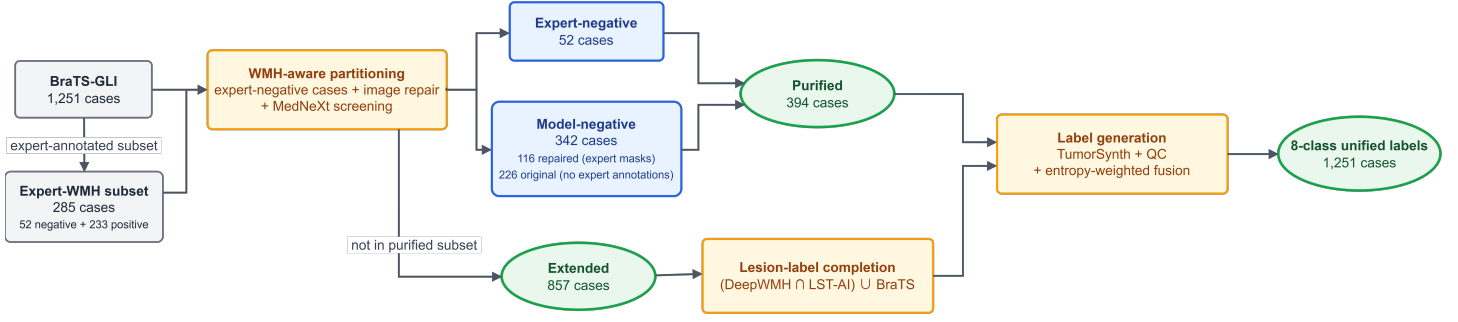}
		\caption{Construction workflow of \resource}
		\label{fig:pipeline}
	\end{figure*}
	
	\subsection{Ethical considerations}
	This resource consists of controlled-access derived labels generated from BraTS de-identified images and labels. It does not add participant recruitment, scanning, clinical-variable collection, or identity-identifying information, so this derived-label organization work does not require a new ethics approval or exemption process. Informed consent, ethics approval, de-identification, data-release authorization, and the final release decision for the original BraTS data are handled by the upstream BraTS/Synapse data workflow. This resource inherits the upstream access restrictions and use conditions and does not replace or bypass the BraTS 2023-GLI data-access process.
	
	This resource is limited to scientific research that complies with laws, research ethics, and upstream BraTS/Synapse terms. It is not intended for direct clinical diagnosis, treatment decisions, or individual risk assessment. Users must not attempt re-identification and must not share downloaded labels or derived metadata with individuals who have not been approved for access to this resource. If suspected data leakage, violation of access conditions, derived-label errors, or version-record errors are found, users should contact the resource contact for this paper, \texttt{xxy200200@stu.xjtu.edu.cn}; issues involving original BraTS data, upstream consent, approval, or licensing terms should be handled by the official BraTS or Synapse management workflow.

	\section{Methods}
	\subsection{Data details}
	The data source is a BraTS 2023-GLI adult brain glioma training-set cohort with 1251 cases. Each case contains T1, T1ce, T2, and T2-FLAIR MRI provided by the upstream BraTS project; these MRI volumes are not redistributed in this resource. The release is labels-only and contains 1367 NIfTI label files: 1251 unified segmentation labels and 116 image-repair labels. The derived unified labels merge tumor and comorbid lesions into Lesion and add six healthy-tissue classes.
	
	Upstream preprocessing information for the released BraTS images is provided in the official CaPTk BraTS Pre-processing Pipeline documentation: \url{https://cbica.github.io/CaPTk/preprocessing_brats.html}. This paper generates derived labels based on the official released results and does not redefine or modify the upstream preprocessing workflow.
	
	The release directory preserves the BraTS-compatible per-case folder layout. Cases requiring upstream MRI repair additionally provide repair labels under \path{repair_labels/modified/}.

	Public metadata files summarize released case/session identifiers, subset and image-repair status, case paths and upstream references, label provenance, QC, license/access terms, size and SHA-256 integrity records, and internal ID mappings used for probability-map generation and label fusion. The released case/session identifier is not claimed to be a verified unique natural-person identifier, and internal IDs are not treated as participant identifiers. Age, sex, scanner, magnetic-field strength, scan protocol, and clinical or molecular variables are controlled by the upstream BraTS workflow; fields not officially released are recorded as unavailable and are not inferred by this resource.
	
	Anonymization, informed consent, original ethics approval, and release decisions for the original data are handled by the upstream BraTS/Synapse workflow; this resource inherits the upstream de-identification status, access restrictions, and use conditions. Case-level provenance records distinguish upstream BraTS tumor labels, expert-negative WMH conclusions and repair-label WMH components from Rudie et al.~\citep{rudie2019multi}, and automatic labels or probability maps from DeepWMH, LST-AI, and TumorSynth. 
	
	The same paired-label comparison also provides a quantitative view of the released label content (Table~\ref{tab:value-added}). The original BraTS-GLI expert lesion foreground is fully retained in the final Lesion class. The added Lesion voxels are concentrated in the extended subset by construction, whereas the six healthy-tissue classes are present in all cases. This result clarifies the resource contribution at two scales. At the lesion scale, the release adds explicit comorbid-lesion constraints in 857 cases. At the whole-label scale, the release changes BraTS-GLI from a lesion-only target into a dense anatomy-lesion supervision target with auditable provenance.

	\begin{table*}[!t]
		\centering
		\caption{Quantitative label content added relative to the original BraTS-GLI expert lesion masks.}
		\label{tab:value-added}
		\footnotesize
		\setlength{\tabcolsep}{1.5pt}
		\renewcommand{\arraystretch}{1.12}
		\renewcommand{\tabularxcolumn}[1]{m{#1}}
		\begin{tabularx}{\textwidth}{@{}>{\centering\arraybackslash}m{0.15\textwidth}*{8}{>{\centering\arraybackslash}X}@{}}
			\toprule
			Metric & GM & BG & WM & \makecell[c]{Lesion outside\\expert mask} & Ven & Cer & BS & \makecell[c]{All new\\foreground} \\
			\midrule
			Cases with new voxels & 1251 & 1251 & 1251 & 857 & 1251 & 1251 & 1251 & 1251 \\
			New voxels (million) & 685.57 & 56.86 & 518.34 & 2.20 & 33.06 & 176.33 & 41.28 & 1513.64 \\
			\makecell[c]{Share of released\\foreground (\%)} & 41.96 & 3.48 & 31.73 & 0.13 & 2.02 & 10.79 & 2.53 & 92.65 \\
			\bottomrule
		\end{tabularx}
	\end{table*}

	\subsection{Methods used for the data creation}
	The overall construction workflow is shown in Figure~\ref{fig:pipeline}. The following text presents three steps: purified subset construction, extended subset label completion, and unified anatomy-lesion labels. These three steps serve the same resource objective: to organize the originally separated tumor subregions, coexisting WMH information, and healthy tissue structures in BraTS-GLI into a single label space, while retaining case stratification and quality status so that downstream research can directly compare data quality and joint supervision.

	\paragraph{Purified subset construction.}
	This study introduces expert WMH annotations for 285 BraTS cases provided by Rudie et al.~\citep{rudie2019multi}; these cases come from the BraTS 2018 training set and fall within the BraTS 2023-GLI training set used in this paper. We used the original T1 and FLAIR images and expert lesion annotations of these 285 cases to train MedNeXt~\citep{isensee2021nnunet,roy2023mednext} in a five-fold configuration, enabling the model to segment tumor regions and coexisting WMH separately. T1 and FLAIR were selected because the target abnormality to be screened out in this study is WMH: FLAIR presents white matter hyperintensities, and T1 provides complementary information.
	
	For the 233 cases with WMH in the expert annotations, NiftySeg~\citep{cardoso2012niftyseg} was used to perform in-place filling of the WMH regions, replacing them with surrounding normal-tissue signal. The trained MedNeXt then screened all candidate cases without an expert-negative conclusion and considered only predicted WMH outside the existing tumor masks. Screening identified 342 model-negative cases. Of these, 116 came from positive cases with expert WMH annotations and had already been repaired; the other 226 came from cases without expert WMH annotations. These 342 cases, together with the 52 expert-negative cases, formed the 394-case purified subset. For the 116 repaired cases, the repaired images replaced the corresponding original images in the cohort and were not counted as duplicate cases. The remaining 857 cases forming the extended subset retained their original imaging status.
	
	\paragraph{Extended subset label completion.}
	For the non-purified 857-case extended subset, the original four-modal images remain unchanged. This paper uses DeepWMH~\citep{li2024deepwmh} and LST-AI~\citep{wiltgen2024lstai} to generate candidate masks for coexisting abnormalities, and takes their intersection to reduce false positives from any single tool. This intersection mask is united with the existing BraTS lesion mask to form the unified Lesion class.
	
	\paragraph{Unified anatomy-lesion labels.}
	To obtain healthy-tissue labels, TumorSynth~\citep{wu2026tumorsynth} was run separately on the four MRI modalities to obtain voxel-level softmax probability maps for each modality. The original outputs were aggregated into seven foreground classes through a mapping table: GM, BG, WM, Lesion, Ven, Cer, and BS. We introduced automatic outlier detection based on the interquartile range (IQR): at the whole-dataset level, the total number of foreground voxels predicted by each modality was counted, and predictions below the lower outlier bound $Q_1-1.5\times\mathrm{IQR}$ were marked as outliers. After manual review, low-quality modality probability maps that caused missing foreground were removed to prevent noisy predictions from entering subsequent fusion.

	Let the lesion hard constraint at voxel $\mathbf{v}$ be $M(\mathbf{v})$, and let the lesion class index be $c_l$. For the probability map of valid modality $m$, the correction is:
	\begin{equation}
		P'_m(c \mid \mathbf{v}) =
		\begin{cases}
			1, & M(\mathbf{v}) > 0 \text{ and } c = c_l, \\
			0, & M(\mathbf{v}) > 0 \text{ and } c \ne c_l, \\
			0, & M(\mathbf{v}) = 0 \text{ and } c = c_l, \\
			\dfrac{P_m(c \mid \mathbf{v})}{\sum_{k \ne c_l} P_m(k \mid \mathbf{v})}, &
			M(\mathbf{v}) = 0 \text{ and } c \ne c_l .
		\end{cases}
		\label{eq:prob-correction}
	\end{equation}
	After the lesion channel is removed outside the lesion region, the corrected probabilities are renormalized. The voxel-wise uncertainty and adaptive weight of each valid modality are:
	\begin{align}
		H_m(\mathbf{v}) &=
		-\sum_c P'_m(c \mid \mathbf{v}) \log P'_m(c \mid \mathbf{v}), \label{eq:entropy}\\
		W_m(\mathbf{v}) &=
		\frac{\exp[-H_m(\mathbf{v})]}{\sum_{j=1}^{N}\exp[-H_j(\mathbf{v})]},
		\label{eq:entropy-weight}
	\end{align}
	where $N$ denotes the number of valid modalities retained after quality control, and entropy is computed using the conventional definition $0\log0=0$. The final fused probability and integer label are:
	\begin{equation}
		L_{\mathrm{fused}}(\mathbf{v}) =
		\arg\max_c
		\sum_{m=1}^{N} W_m(\mathbf{v})P'_m(c \mid \mathbf{v}).
		\label{eq:fused-label}
	\end{equation}
	The labels generated by Eqs.~\eqref{eq:prob-correction} to \eqref{eq:fused-label} fix the unified lesion constraint and preserve high-confidence predictions of healthy anatomical structures from each modality.
	\begin{table*}[!t]
		\centering
		\caption{Comparison of joint segmentation performance}
		\label{tab:cross-lesion}
		\footnotesize
		\setlength{\tabcolsep}{3pt}
		\resizebox{\textwidth}{!}{%
			\begin{tabular}{cclccccccc}
				\toprule
				\multicolumn{1}{c}{Dataset} & \multicolumn{1}{c}{Method} & \multicolumn{1}{c}{Metric} & GM & BG & WM & Lesion & Ven & Cer & BS \\
				\midrule
				\multirow{4}{*}{\makecell[c]{GLI\\(in-domain)}}
				& \multirow{2}{*}{Baseline}
				& DSC$\uparrow$ & 96.0 (1.9) & 95.2 (3.9) & 95.9 (2.2) & 90.5 (9.8) & 94.8 (3.8) & 98.4 (2.9) & 97.2 (3.4) \\
				&
				& HD95$\downarrow$ & 1.07 (0.29) & 1.21 (0.64) & 1.10 (0.36) & \textbf{15.71 (18.47)} & 1.25 (0.83) & 1.21 (1.89) & 1.22 (1.70) \\
				& \multirow{2}{*}{Joint-Baseline}
				& DSC$\uparrow$ & 96.3 (1.6) & 95.9 (3.1) & 96.2 (1.9) & \textbf{90.8 (10.3)} & 95.7 (2.8) & 98.5 (2.8) & 97.6 (3.1) \\
				&
				& HD95$\downarrow$ & 1.05 (0.23) & 1.11 (0.43) & 1.09 (0.30) & 18.90 (30.06) & 1.11 (0.41) & 1.19 (1.88) & 1.21 (1.70) \\
				\midrule
				\multirow{4}{*}{\makecell[c]{WMH\\(out-of-domain)}}
				& \multirow{2}{*}{Baseline}
				& DSC$\uparrow$ & 79.1 (2.1) & 88.3 (2.0) & 89.8 (2.7) & \textbf{31.1 (27.5)} & 89.1 (3.5) & 91.4 (1.5) & 87.4 (2.1) \\
				&
				& HD95$\downarrow$ & 6.55 (1.34) & 1.67 (0.32) & 1.65 (0.35) & \textbf{140.40 (150.46)} & 1.84 (0.48) & 4.84 (0.99) & 2.31 (0.49) \\
				& \multirow{2}{*}{Joint-Baseline}
				& DSC$\uparrow$ & 79.3 (2.1) & 88.4 (2.0) & 89.4 (3.2) & 4.3 (12.9) & 89.6 (3.4) & 91.3 (1.6) & 87.1 (2.2) \\
				&
				& HD95$\downarrow$ & 6.50 (1.33) & 1.68 (0.32) & 1.92 (0.64) & 297.46 (132.37) & 1.77 (0.48) & 4.84 (1.01) & 2.34 (0.51) \\
				\bottomrule
			\end{tabular}
		}
	\end{table*}
	
	\section{Validation}
	\subsection{Description of approaches ensuring quality of data}
	Quality control corresponds to four levels of the data-creation workflow and release integrity. First, expert-negative samples in the purified subset come from cases with strictly zero WMH volume among the 285 expert WMH annotations; the remaining purified cases were selected as high-confidence WMH-negative samples by five-fold MedNeXt trained from the expert annotations. Second, the additional lesion constraints for the extended subset use the intersection of DeepWMH and LST-AI and are united with existing BraTS lesion masks, reducing false positives from a single tool while retaining the original target lesions. Third, the four-modal TumorSynth probability maps undergo IQR outlier detection based on total foreground voxel counts, and outlier modalities are removed after manual review before entropy-weighted fusion. Fourth, after freezing the release tree, labels-only case-level and file-level integrity checks were completed: case counts and subset counts, unified-label and repair-label paths, file roles, file sizes, SHA-256 digests, access conditions, and manifest alignment were checked. A total of 1251/1251 cases passed and 0 failed; all 1367 NIfTI label files entered the release manifest, including 1251 unified segmentation labels and 116 image-repair labels, and the number of redistributed MRI images was 0.
	
	\begin{figure*}[!t]
		\centering
		\setlength{\tabcolsep}{3pt}
		\renewcommand{\arraystretch}{0.92}
		\begin{tabular}{@{}>{\centering\arraybackslash}m{0.075\textwidth}@{\hspace{0.35em}}
				*{5}{>{\centering\arraybackslash}m{0.155\textwidth}}@{}}
			& \textbf{T1} & \textbf{FLAIR} & \textbf{GT} & \textbf{Baseline} & \textbf{Joint-Baseline} \\
			
			\makecell[c]{\textbf{GLI}\\[-1pt]\scriptsize in-domain}
			& \includegraphics[width=\linewidth]{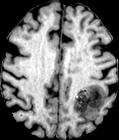}
			& \includegraphics[width=\linewidth]{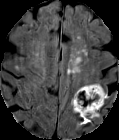}
			& \includegraphics[width=\linewidth]{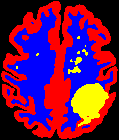}
			& \includegraphics[width=\linewidth]{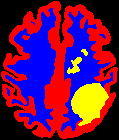}
			& \includegraphics[width=\linewidth]{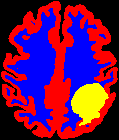} \\
			
			\makecell[c]{\textbf{WMH}\\[-1pt]\scriptsize out-of-\\[-1pt]\scriptsize domain}
			& \includegraphics[width=\linewidth]{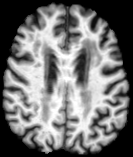}
			& \includegraphics[width=\linewidth]{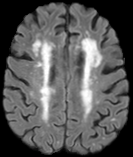}
			& \includegraphics[width=\linewidth]{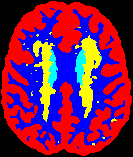}
			& \includegraphics[width=\linewidth]{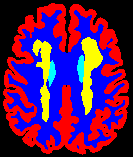}
			& \includegraphics[width=\linewidth]{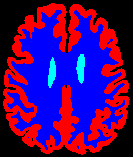}
		\end{tabular}
		\caption{Visualization analysis of in-domain and out-of-domain joint segmentation}
		\label{fig:validation-example}
	\end{figure*}

	\subsection{Controlled ML validation}
	This paper uses a controlled ML use case to answer a resource-level question: in a unified anatomy-lesion label space, why is it necessary to construct and release a purified subset separately, rather than relying only on more noisy training cases? All experiments use MedNeXt-M~\citep{isensee2021nnunet,roy2023mednext} with default settings: 3$\times$3$\times$3 kernel, five-fold cross-validation, AdamW optimizer, initial learning rate $1\times10^{-3}$, weight decay $3\times10^{-5}$, 1000 epochs, and batch size 2. Baseline was trained only on the 394-case purified subset. The 1000 training cases for Joint-Baseline consisted of the 394-case purified subset and 606 additional cases randomly selected from the extended subset, and the latter did not receive additional comorbid-lesion labels. Both experiments used T1 and FLAIR inputs.

	For each foreground class, this paper uses the Dice similarity coefficient (DSC) to measure the overlap between the predicted mask and the reference mask. When both the predicted mask and the reference mask are empty, DSC is recorded as 1.0. HD95 is the 95th percentile of the surface distance between the predicted boundary and the reference boundary; smaller values indicate closer boundaries. It is recorded as 0.0 when both masks are empty. If only one side is empty, if the abnormal prediction fills the whole image, or if distance computation fails or exceeds the threshold, the maximum distance penalty value 374.0 is assigned. Metrics in the table are first computed within each case and then reported as case-level mean (standard deviation).

	GLI in-domain testing used the remaining 251 cases. External WMH testing used 170 images from the MICCAI 2017 WMH dataset, which originally provides T1, FLAIR, and lesion masks~\citep{kuijf2019wmh}. Following the BraTS processing workflow, this paper used HD-BET for skull stripping~\citep{isensee2019hdbet}, used ANTs~\citep{tustison2021antsx} to rigidly register T1 to SRI24~\citep{rohlfing2010sri24}, and applied the same transform to FLAIR and lesion labels. Healthy-tissue labels were generated by WMH-SynthSeg~\citep{laso2024wmhsynthseg} and mapped to the same seven foreground classes as GLI. The same lesion hard constraint and entropy-weighted fusion rules as in the GLI data were then used to generate the joint reference labels. This external validation set is not part of the GLI resource released in this paper.

	\noindent Table~\ref{tab:cross-lesion} shows that, in GLI in-domain evaluation, the healthy-tissue DSC and HD95 of Baseline and Joint-Baseline are close, indicating that a smaller amount of high-quality data can still maintain stable healthy-brain-tissue segmentation ability. The lesion DSC of Joint-Baseline is only slightly higher than that of Baseline, but its HD95 is larger and has a higher standard deviation, indicating that the two models have similar overlap performance for the main lesion, while Joint-Baseline is more likely to miss WMH that is far from the main lesion and smaller in volume, thereby increasing distance error. In external zero-shot WMH evaluation, the two models still maintain similar results for healthy-tissue classes, indicating that model performance on out-of-domain healthy anatomical-structure segmentation is preserved, whereas Baseline has clearly higher lesion DSC and lower HD95 than Joint-Baseline. This result explains the resource value of the purified subset: it provides a WMH-aware cleaned training reference for the joint label space, allowing researchers to distinguish the effects of data quality and sample size on sensitivity to coexisting lesions. The relatively low overall out-of-domain GM metric mainly comes from differences in reference-label generation protocols: healthy tissues in the GLI training labels are generated by TumorSynth, whereas WMH test labels are generated by WMH-SynthSeg. Their gray-matter boundary definitions are not fully consistent, so this phenomenon should not be interpreted as a failure of healthy-tissue segmentation ability. The qualitative results in Figure~\ref{fig:validation-example} are consistent with the table: in the GLI sample, both models cover the main tumor region, and the differences mainly appear in the segmentation of implicit WMH; in the WMH sample, Baseline detects more WMH regions aligned with GT, whereas Joint-Baseline shows no response to WMH lesions.
	
	\FloatBarrier
	\section{Conflicts of Interest}
	The authors declare that they have no conflicts of interest.
	
	\section{Acknowledgements}
	The authors thank the organizers of the Brain Tumor Segmentation (BraTS) Challenge and the data-providing institutions for establishing and releasing de-identified brain MRI research resources. Data used in this publication were obtained as part of the Brain Tumor Segmentation (BraTS) Challenge project through Synapse ID: syn51156910. The authors also thank Rudie et al. for the publicly available expert annotation materials that supported the construction of this derived label resource. This work was supported in part by the National Natural Science Foundation of China (Nos. T2522028 and 12326616), Natural Science Basic Research Program of Shaanxi (No. 2024JC-TBZC-09), and Shaanxi Provincial Key Industrial Innovation Chain Project (No. 2024SF-ZDCYL-02-10).
	
	\begingroup
	% Restore BibTeX DOI printing; melba.cls uses \doi for article metadata.
	\renewcommand{\doi}{doi: \begingroup\urlstyle{rm}\Url}
	\bibliography{sample}
	\endgroup
	
\end{document}